\documentclass[10pt,journal,compsoc]{styles/IEEEtran}

\ifCLASSINFOpdf
  \usepackage[pdftex]{graphicx}
\else
   \usepackage[dvips]{graphicx}
\fi

\ifCLASSOPTIONcompsoc
  \usepackage[nocompress]{cite}
\else
  \usepackage{cite}
\fi

\ifCLASSINFOpdf
\else
\fi

\usepackage{url}

\usepackage{setspace}
\usepackage{algorithm}
\usepackage{algorithmic}

\newcommand{\halfpagewidth}{7.5cm}
\newcommand{\bibliographydir}{./bib}
\graphicspath{{./figures/}}

\def\MyTitle{Fast image-based obstacle detection from unmanned surface vehicles}

\begin{document}

\title{\MyTitle}

\author{Matej~Kristan,~\IEEEmembership{Member,~IEEE,}
        Vildana Suli{\' c} Kenk,
        Stanislav Kova{\v c}i{\v c},~\IEEEmembership{Member,~IEEE,}
        Janez Per{\v s},~\IEEEmembership{Member,~IEEE,}
\thanks{This work was supported in part by the Slovenian research agency programs \mbox{P2-0214}, \mbox{P2-0094}, and projects J2-4284, J2-3607, J2-2221. We also thank HarphaSea d.o.o. for their hardware used to capture the dataset.}
\thanks{M. Kristan is with the Faculty of Computer and Information Science and with the Faculty
of Electrical Engineering, University of Ljubljana, Slovenia,
e-mail: matej.kristan@fri.uni-lj.si.}
\thanks{V. Suli{\v c}, S. Kova{\v c}i{\v c} and J. Per{\v s} are with the Faculty
of Electrical Engineering, University of Ljubljana, Slovenia.}}

\markboth{Draft, submitted to a journal }
{Kristan, Suli{\v c}, Kova{\v c}i{\v c}, Per{\v s}, \MyTitle}

\IEEEtitleabstractindextext{%
\begin{abstract}
Obstacle detection plays an important role in unmanned surface vehicles (USV). The USVs operate in highly diverse environments in which an obstacle may be a floating piece of wood, a scuba diver, a pier, or a part of a shoreline, which presents a significant challenge to continuous detection from images taken onboard. This paper addresses the problem of online detection by constrained unsupervised segmentation. To this end, a new graphical model is proposed that affords a fast and continuous obstacle image-map estimation from a single video stream captured onboard a USV. The model accounts for the semantic structure of marine environment as observed from USV by imposing weak structural constraints. A Markov random field framework is adopted and a highly efficient algorithm for simultaneous optimization of model parameters and segmentation mask estimation is derived. Our approach does not require computationally intensive extraction of texture features and comfortably runs in real-time. The algorithm is tested on a new, challenging, dataset for segmentation and obstacle detection in marine environments, which is the largest annotated dataset of its kind. Results on this dataset show that our model outperforms the related approaches, while requiring a fraction of computational effort.
\end{abstract}

}

\maketitle

\IEEEdisplaynontitleabstractindextext
\IEEEpeerreviewmaketitle

\IEEEraisesectionheading{\section{Introduction}\label{sec:introduction}}

\begin{figure}
        \centering
            \includegraphics[width=8.5cm]{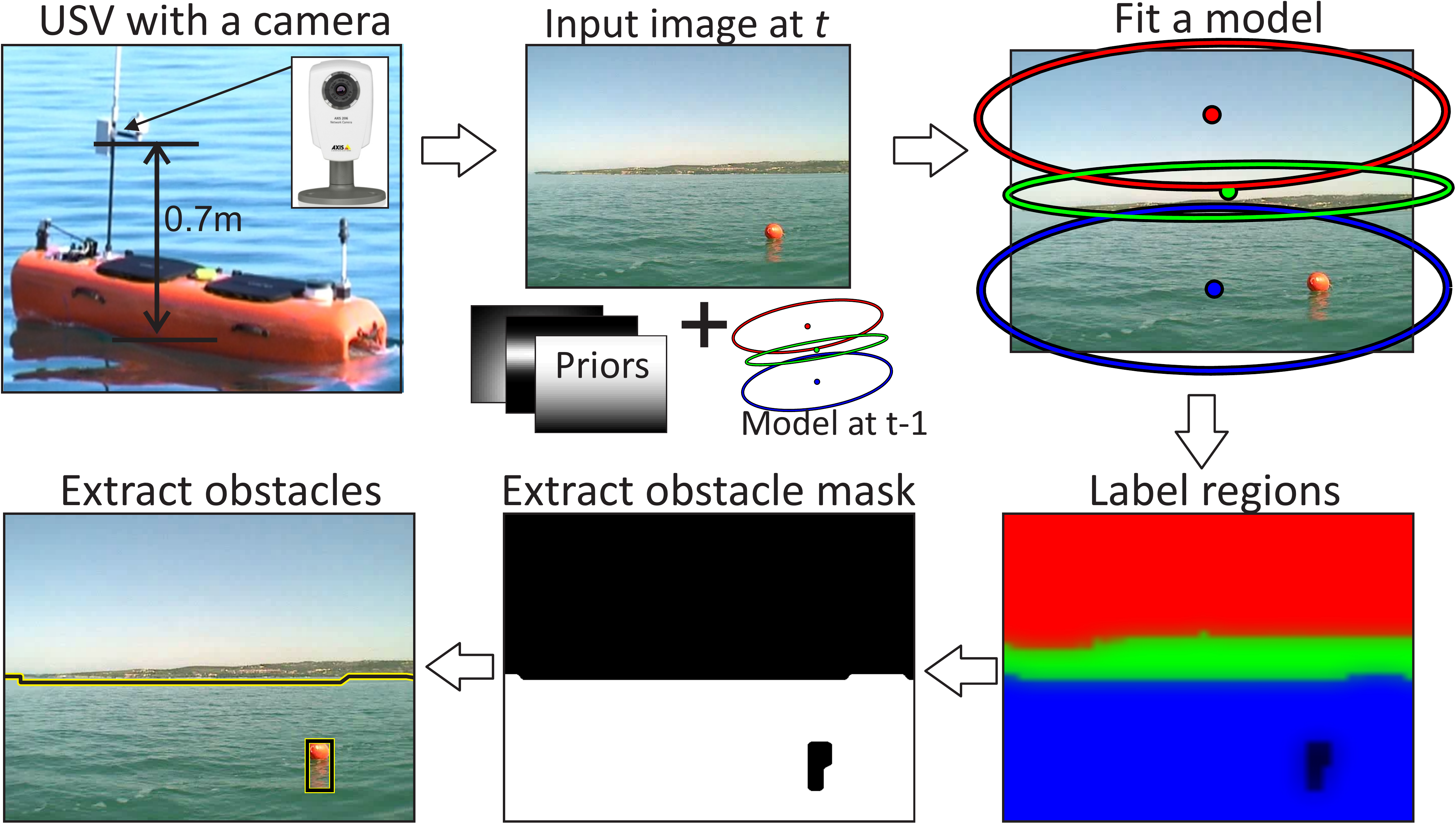}
        \caption{\label{fig:ourapproach} Our approach to obstacle image-map estimation.}
\end{figure}

\IEEEPARstart{O}{bstacle} detection is of central importance for lower-end small unmanned surface vehicles (USV) used for patrolling coastal waters (see Figure~\ref{fig:ourapproach}). Such vehicles are typically used in perimeter surveillance, in which the USV travels along a pre-planned path. To quickly and efficiently respond to the challenges from highly dynamic environment, the USV requires an onboard logic to observe the surrounding, detect potentially dangerous situations, and apply proper route modifications. An important feature of such vessel is the ability to detect an obstacle at sufficient distance and react by replanning its path to avoid collision. The primary type of obstacle in this case is the shoreline itself, which can be avoided to some extent (although not fully) by the use of detailed maps and the satellite navigation. Indeed, Heidarsson and Sukhatme~\cite{heidarsson2011image} proposed an approach that utilizes an overhead image of the area obtained from Google maps to construct a map of static obstacles. But such an approach cannot handle a more difficult class of dynamic obstacles that do not appear in the map (e.g., boats, buys and swimmers).

A small USV requires ability to detect near-by and distant obstacles. The detection should not be constrained to objects that stand out from the water, but should also detect flat objects, like debris or emerging scuba divers, etc. Operation in shallow waters and marinas constrains the size of USV and prevents the use of additional stabilizers. This puts further constraints on the weight, power consumption, types of sensors and their placement. Cameras are therefore becoming attractive sensors for use in low-end USVs due to their cost-, weight- and power-efficiency and a large field of view coverage. This presents a challenge for development of highly efficient computer vision algorithms tailored for obstacle detection in a challenging environments that the small USVs face. In this paper we address this challenge by proposing a segmentation-based algorithm for obstacle-map estimation that is derived from optimizing a new well-defined graphical model and runs at over 70fps in Matlab on a single core machine.

\subsection{Related work}

The problem of obstacle detection has been explicitly or implicitly addressed previously in the field of unmanned ground vehicles (UGV). In a trail-following application Rasmussen et al.~\cite{Rasmussen2010} use an omnidirectional camera to detect trail as a region that is most contrasted to its surrounding, however, dynamic obstacles are not addressed. Several works, e.g., Montemerlo et al.~\cite{Montemerlo2006} and Dahlkamp et al.~\cite{Dahlkamp2006}, address the problem of low-proximity road detection with laser scanners by bootstrapping color segmentation with the laser output. The proximal road points are detected by laser, projected to camera and used to learn a Gaussian mixture model which is in turn used to segment the rest of the image captured by the camera. Combined with horizon detection of Ettinger et al.~\cite{Ettinger2003}, this approach significantly increases the distance at which the obstacles on the road can be detected. Alternatively, Lu and Rasmussen~\cite{Lu2012} casted the obstacle detection as a labelling task in which they employ a bank of pre-trained classifiers to 3D point clouds and a Markov random field to account for the spatial smoothness of the labelling.

Most UGV approaches for obstacle detection explicitly or implicitly rely on ground plane estimation from range sensors and are not directly applicable to aquatic environments encountered by USV. Rankin et al.~\cite{Rankin2010} propose a specific body-of-water detector in wide open areas from a UGV using a monocular color camera. Their detector assumes that, in an undisturbed water surface, a change in saturation-to-brightness ratio across a water body from the leading to trailing edge is uniform and distinct from other terrain types. They apply several ad-hoc processing steps to gradually grow the water regions for the initial candidates and apply a sequence of pre-set thresholds to remove spurious false detections of water pools. However, their method is based on the undisturbed water surface assumptions, which is violated in coastal and open water applications. Scherer et al.~\cite{Scherer2012} propose a water detection algorithm using a stereo bumblebee camera, IMU/GPS and rotating laser scanner for navigation on a river. Their system extracts color and texture features over blocks of pixels and eliminates the sky region using a pre-trained classifier. A horizon line, obtained from the onboard IMU, is then projected into the image to obtain samples for learning a color distribution of the regions below and above horizon, respectively. Using these distributions, the image is segmented and results of the segmentation are used in turn, after additional postprocessing steps, to train a classifier. The trained classifier is fused with a classifier from the previous frames and applied to the blocks of pixels to detect the water region. This system relies heavily on the quality of hardware-based horizon estimation, accuracy of pre-trained sky detector and the postprocessing steps. The authors report that the vision-based segmentation is not processed onboard, but requires special computing hardware, which makes it below a realtime segmentation at constrained processing power typical for small USVs.

Some of the standard range sensor modalities for autonomous navigation in maritime environments include radar~\cite{Onunka2010}, sonar~\cite{heidarsson2011} and ladar~\cite{Rankin2010}. Range scanners are known to poorly discriminate between water and land in the far field~\cite{Elkins2010}, suffer from angular resolution and scanning rate limitations, and poorly perform when the beam's incidence angle is not oblique with respect to the water surface~\cite{Hong02fusingladar,Santana2012}. Several researchers have thus resorted to cameras, e.g.,~\cite{socek2005,fefilatyev2008,Rankin2010,wang2011,Huntsberger2011,Santana2012}, for obstacle and moving object detection instead. To detect dynamic objects in harbor, Socek et al.~\cite{socek2005} assume a static camera and apply background subtraction combined with motion cues. However, background subtraction cannot be applied to a highly dynamic scenes encountered on a moving USV. Huntsberger et al.~\cite{Huntsberger2011} attempt to address this issue using stereo systems, but require large baseline rigs that are less appropriate for small vessels due to increased instability and limit processing of near-field regions. Santana et al.~\cite{Santana2012} apply fusion of Lukas Kanade local trackers with color oversegmentation and a sequence of k-means clusterings on texture features to detect water regions in videos. Alternatively, Fefilatyev and Goldgof~\cite{fefilatyev2008} and Wang et al.\cite{wang2011} apply a low-power solution using a monocular camera for obstacle detection. They
first detect the horizon line and then search for a potential obstacle in the region below the horizon. A fundamental drawback of these approaches is that they approximate the edge of water by a horizon line and cannot handle situations in coastal waters, close to the shoreline or in marina. At that point, the edge of water does not correspond to the horizon anymore and can be no longer modeled as a straight line. Such cases call for more general segmentation approaches.

Many unsupervised segmentation approaches have been proposed in literature. Khan and Shah~\cite{Khan2001} use optical flow, color and spatial coordinates to construct features which are used in single Gaussians to segment a moving object in video. Nguyen and Wu~\cite{Nguyen2013} propose Student-t mixture models for robustifying segmentation. Improved segmentation can be achieved by applying Bayesian regularization scheme in Gaussian mixture models, however, care has to be taken at initialization~\cite{Nasios2006}. Felzenswalb and Huttenlocher~\cite{Felzenszwalb2004} have proposed a graph-theoretic clustering to perform segmentation of color images into visually-coherent regions. The assumption that the neighboring pixels likely belong to the same class is formally addressed in the context of Markov random fields (MRF)~\cite{Besag1986,boykov2006graph}. By constraining the solutions of the segmentations to mimic high-level semantics of urban scenes, Felzenszwalb and Veksler~\cite{Felzenszwalb2010a} proposed a three-strip segmentation algorithm that can be implemented by a dynamic program. Wojek and Schiele~\cite{Wojek2008} have extended the conditional random fields with dynamic models and perform the inference for object detection and labeling jointly in videos. The random field frameworks~\cite{lafferty2001conditional} have proven quite successful for addressing the semantic labeling tasks and recently Kontschieder et al.~\cite{kontschieder2011structured} have shown that structural priors between classes further improve the labeling. Alternative schemes that avoid applying a MRF to enforce spatial consistency have been proposed, e.g., Chen et al.~\cite{Chen2004a} and Nguyen et al.~\cite{Nguyen2012}. The approaches like Wojek et al.~\cite{Wojek2008} use high-dimensional features composed of color and texture at multiple scales and object-class specific detectors to segment the images and detect the objects of interest. In our scenarios, the possible types of dynamic obstacles are unknown and vary significantly in appearance. Thus object-class specific detectors are not suitable. Several bottom-up graph-theoretic approaches have been proposed for unsupervised segmentation, e.g.,~\cite{Makrogiannis2005,Tao2007,Alpert2012,Li2012}. Recently, Alpert et al.~\cite{Alpert2012} have proposed an approach that starts from a pixel level and gradually constructs visually-homogenous regions by agglomerative clustering. They achieved impressive results on a segmentation dataset in which an object was occupying a significant portion of an image. Unfortunately, since their algorithm incrementally merges regions, it is too slow for online application even at moderate image sizes. An alternative to starting the segmentation from pixel level is to start from an oversegmented image such that pixels are grouped into superpixels~\cite{Ren2003}. Lu et al.~\cite{Lu2013} apply spectral clustering to an affinity graph induced over a superpixelated image. Li et al.~\cite{Li2012} have proposed a segmentation algorithm that uses multiple superpixel oversegmentations and merges their result by a bipartite graph partitioning to achieve state-of-the-art results on a standard segmentation dataset. However, no prior information is provided to favor certain types of segmentations in specific scenes.

\subsection{Our approach}

\begin{figure}[h!]
        \centering
            \includegraphics[width=\halfpagewidth]{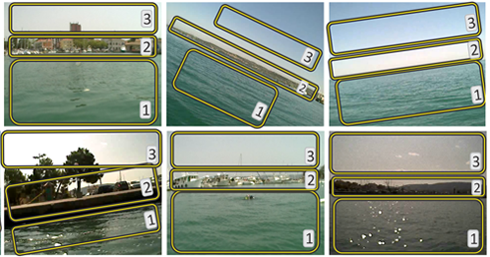}
        \caption{\label{fig:whatboatsees} Images captured from the USV split into three semantically different regions.}
\end{figure}

We pursue a solution for obstacle detection that is based on concepts of image segmentation with weak semantic priors on the expected scene composition. Figure~\ref{fig:whatboatsees} shows typical images captured from a USV. While the images significantly vary in appearance, we observe that each image can be split into three semantic regions roughly stacked one above the other, implying a structural relation between the regions. The bottom region represents the water, while the top region represents the sky. The middle component can represent either land, parked boats a haze above horizon or a mixture of these.

Our \textbf{main contribution} is a graphical model for structurally-constrained semantic segmentation with application to USV obstacle-map estimation. The generative model assumes a mixture model with three Gaussian components for the dominant three image regions and a uniform component for explaining the outliers, which may constitute an obstacle in the water. Weak priors are assumed on the mixture parameters and a MRF is placed over the prior as well as posterior pixel-class distributions to favor smooth segmentations. We derive an EM algorithm for the proposed model and show that the resulting optimization achieves a fast convergence at a low computational cost, without resorting to a specialized hardware. A similar graphical model was proposed by Diplaros et al.~\cite{Diplaros2007}, but their model requires a manually set variable, does not apply priors and is not derived from a single density function. Our model is applied to obstacle image-map estimation in USVs. The proposed model acts directly on color image and does not require expensive extraction of texture-based features. Combined with efficient optimization, this results in realtime segmentation and obstacle-map estimation (several-fold faster than the camera frame rate). Our approach is outlined in Figure~\ref{fig:ourapproach}. The semantic model is fitted to the input image, after which each pixel is classified into one of the four classes. All the pixels that do not correspond to the water component are deemed to be a part of an obstacle. Figure~\ref{fig:ourapproach} shows a detection of a dynamic obstacle (buoy) and of a static obstacle (shoreline). Our \textbf{second contribution} is a marine dataset for semantic segmentation and obstacle detection, and the performance evaluation methodology. To our knowledge this will be the largest annotated publicly available marine dataset of its kind up to date.

A preliminary version of our algorithm was presented in Kristan et al.~\cite{Kristan2014a} and is extended in this paper on several levels. Additional discussion and related work is provided. Improved initialization of segmentation model by soft-resets of the parameters is proposed and additional details of the algorithm and the dataset are provided. In particular, the dataset capturing procedure and annotation is discussed and additional statistics of the obstacles in the dataset are provided. The experiments are extended by performance analysis with respect to the color space, the obstacle size and the time-of-day driving conditions. The learning of priors used in our model is discussed in detail and the dataset is extend with training images used for estimating the priors.

Our approach is most closely related to the works in urban-scene parsing by Felzenszwalb and Veksler~\cite{Felzenszwalb2010a} and maritime scene understanding by Fefilatyev and Golggof~\cite{fefilatyev2008}, Wang et al.,~\cite{wang2011} and Scherer et al.~\cite{Scherer2012}. There are notable differences between these approaches and ours. The first difference to~\cite{Felzenszwalb2010a} is that they only address the labeling part of the segmentation problem and require precomputed per-pixel label confidences. The second difference is that their approach produces segmentations with homogenous bottom region, which prevents detection of obstacles without further postprocessing. In contrast, our approach jointly learns the component appearance, estimates the per-pixel class probabilities, and optimizes the segmentation within a single online framework. Furthermore, learning the parameters of~\cite{Felzenszwalb2010a} is not as straightforward. Compared to the related water segmentation algorithms for maritime applications (i.e.,~\cite{fefilatyev2008,wang2011,Scherer2012}), our approach completely avoids the need for a good horizon estimation. Nevertheless, the proposed probabilistic model is general enough to directly incorporate this information if available.

The remainder of the paper is structured as follows. In Section~\ref{sec:segmentationModel} we derive our semantic generative model, in Section~\ref{sec:obstacleDetectionAlgorithm} we present the obstacle detection algorithm, in Section~\ref{sec:implementDetails} we detail the implementation and learning of the priors, in Section~\ref{sec:modddataset} we present the new dataset and the accompanying evaluation protocol, in Section~\ref{sec:experiments} we experimentally analyze the algorithm and draw conclusions in Section~\ref{sec:conclusion}.

\section{The semantic generative model}\label{sec:segmentationModel}

\begin{figure}
        \centering
            \includegraphics[width=6cm]{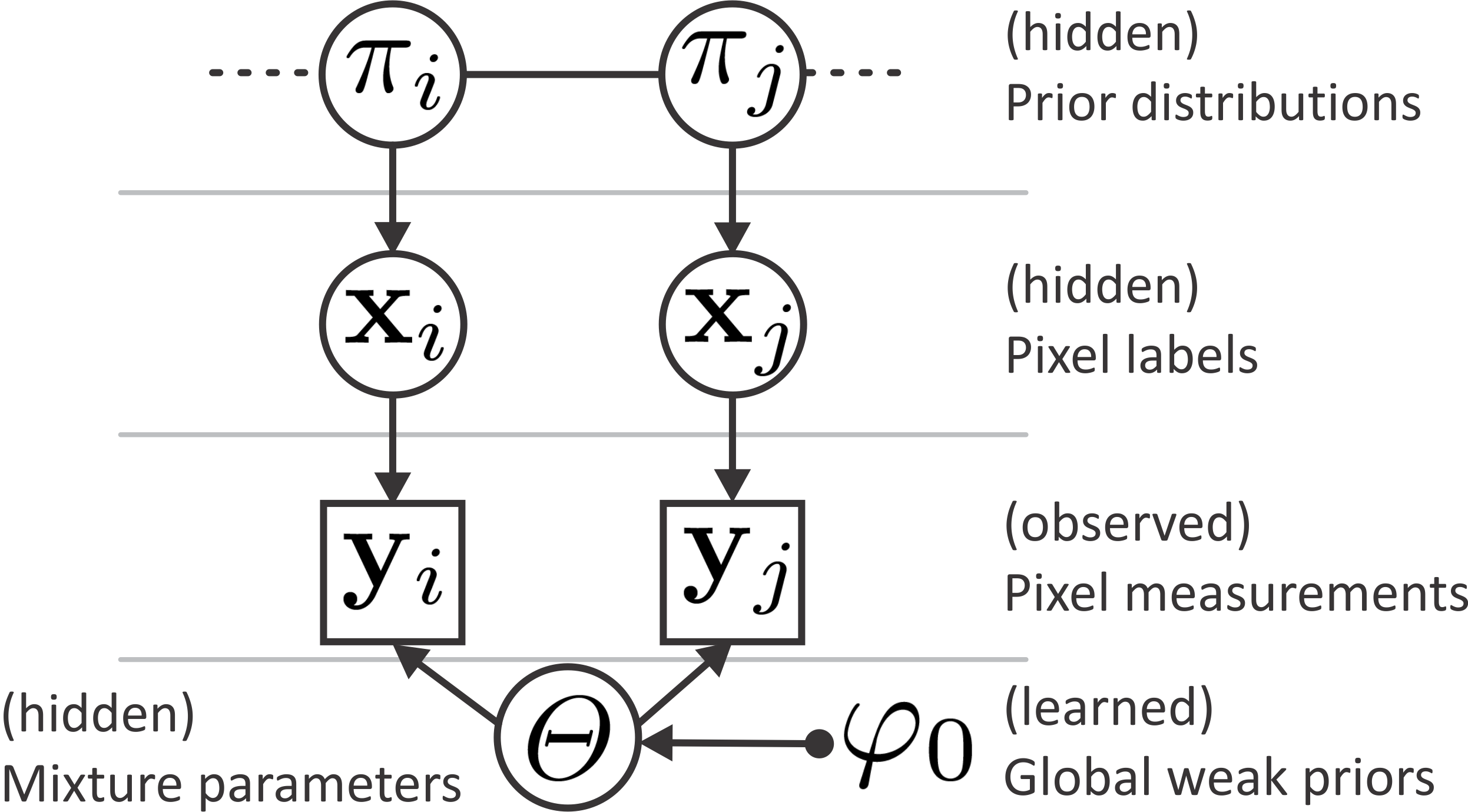}
        \caption{\label{fig:graphicalmodel} The graphical model for semantic segmentation.}
\end{figure}

We consider the image as an array of measured values $\mathbf{Y}= \{ \mathbf{y}_i \}_{i=1:M}$, in which $\mathbf{y}_i \in \mathcal{R}^d$ is a $d$ dimensional measurement, a feature vector, at the $i$-th pixel in an image with $M$ pixels. As we detail in the subsequent sections, the feature vector is composed of pixel's color and image coordinates. The probability of the $i$-th pixel feature vector is modelled as a mixture model with four components -- three Gaussians and a single uniform component:
\begin{eqnarray}\label{eq:measurementmodel}
    p(\mathbf{y}_i | \Theta ) = \sum\limits_{k=1}^3 \phi(\mathbf{y}_i | \mu_k, \Sigma_k) \pi_{ik} + \mathcal{U}(\mathbf{y}_i)\pi_{i4},
\end{eqnarray}
where $\Theta=\{ \mu_k, \Sigma_k \}_{k=1:3}$ are the means and covariances of the Gaussian kernels $\phi(\cdot| \mu, \Sigma)$ and $\mathcal{U}(\cdot)$ is a uniform distribution. The $i$-th pixel label $x_i$ is an unobserved random variable governed by the class prior distribution $\pi_i=[\pi_{i1},\dots,\pi_{il},\dots,\pi_{i4}]$ with $\pi_{il}=p(x_i=l)$. The three Gaussian components represent the three dominant semantic regions in the image, while the uniform component represents the outliers, i.e., pixels that do not likely correspond to any of the three structures. To encourage segmentations into three approximately vertically aligned semantic structures, we define a set of priors $\varphi_0=\{ \mu_{\mu_k}, \Sigma_{\mu_k} \}_{k=1:3}$ for the mean values of the Gaussians, i.e., $p(\Theta | \varphi_0)=\prod\nolimits_{k=1}^3 \phi(\mu_k | \mu_{\mu_k}, \Sigma_{\mu_k})$. To encourage smooth segmentations, the priors $\pi_i$ as well as posteriors over the pixel class labels, are treated as random variables, which form a Markov random field. Imposing the MRF on the priors and posteriors rather than pixel labels allows effectively integrating out the labels, which leads to a well-behaved class of MRFs~\cite{Diplaros2007} that avoid image reconstruction during parameter learning. The resulting graphical model with priors is shown in Figure~\ref{fig:graphicalmodel}.

Let $\pi = \{ \pi_{i} \}_{i=1:M}$ denote the set of priors for all pixels. Following~\cite{Besag1986} we approximate the joint distribution over the priors as $p(\pi)\approx \prod\nolimits_i p(\pi_i | \pi_{N_i})$, and $\pi_{N_i}$ is a mixture distribution over the priors of the $i$-th pixel's neighbors, i.e., $\pi_{N_i}=\sum\nolimits_{j \in N_i , j\neq i} \lambda_{ij} \pi_j$, where $\lambda_{ij}$ are fixed positive weights such that for each $i$-th pixel $\sum\nolimits_j \lambda_{ij}=1$. The potentials in the MRF are defined as
\begin{equation}
    p(\pi_i | \pi_{N_i}) \propto \exp{(-{1 \over 2} E(\pi_i, \pi_{N_i}) )},
\end{equation}
with the exponent defined as
\begin{equation}
   E( \pi_i, \pi_{N_i})= D(\pi_i \parallel \pi_{N_i})+H(\pi_i).
\end{equation}
The term $D(\pi_i \parallel \pi_{N_i})$ is the Kullback-Leibler divergence which penalizes the differences between prior distributions over the neighboring pixels ($\pi_i$ and $\pi_{N_i}$), while the term $H(\pi_i)$ is the entropy defined as
\begin{equation}
    H(\pi_i) = -\sum\limits_{i=1}^4 \pi_{i k} \log\pi_{i k},
\end{equation}
which penalizes uninformative priors $\pi_i$. The joint distribution for the graphical model in Figure~\ref{fig:graphicalmodel} can be written as
\begin{equation}
    p(\mathbf{Y}, \Theta, \pi | \varphi_0) = \prod\limits_{i=1}^M p(\mathbf{y}_i | \Theta, \varphi_0) p(\Theta| \varphi_0) p(\pi_i | \pi_{N_i}).
\end{equation}
Diplaros et al.~\cite{Diplaros2007} argue that improved segmentations can be achieved by also considering an MRF directly on the pixel posterior distributions by treating the posteriors as random variables $\mathbf{P} = \{ \mathbf{p}_i \}_{i=1:M}$, where the components of $\mathbf{p}_i$ are defined as $p_{ik}= p(x_i=k| \Theta, \mathbf{y}_i, \varphi_0)$, computed by Bayes rule from $p(y_i|x_i=k, \Theta)$ and $p(x_i=k)$. We can write the posterior over $\mathbf{P}$ as $p(\mathbf{P} | \mathbf{Y}, \Theta, \pi, \varphi_0) \propto \prod\nolimits_{i=1}^M \exp(-{1 \over 2} E(\mathbf{p}_i,\mathbf{p}_{N_i}))$, where $\mathbf{p}_{N_i}$ is a mixture defined in the same spirit as $\pi_{N_i}$. The joint distribution can now be written as
\begin{eqnarray}\label{eq:cost1}
    p(\mathbf{P}, \mathbf{Y}, \Theta, \pi | \varphi_0) \propto  \exp[\sum\limits_{i=1}^M \log p(\mathbf{y}_i , \Theta|\varphi_0)\nonumber\\  -  {1 \over 2}( E(\pi_i, \pi_{N_i}) + E(\mathbf{p}_i,\mathbf{p}_{N_i}) ) ],
\end{eqnarray}
Due to coupling between $\pi_i$/$\pi_{N_i}$ and $\mathbf{p}_i$/$\mathbf{p}_{N_i}$ the optimization of (\ref{eq:cost1}) is not straightforward. We therefore introduce auxiliary variables $\mathbf{q}_i$ and $\mathbf{s}_i$ and take the logarithm, which results in the following cost function
\begin{equation}\label{eq:cost2}
    F = \sum\limits_{i=1}^M [\log p(\mathbf{y}_i , \Theta|\varphi_0) - \small{\frac{1}{2}}( D( \mathbf{s}_i \| \pi_i \circ \pi_{N_i}) + D( \mathbf{q}_i \| \mathbf{p}_i\circ\mathbf{p}_{N_i}) ) ],
\end{equation}
where $\circ$  is the Hadamard (component-wise) product. Note that when $\mathbf{q}_i\equiv \mathbf{p}_i$ and $\mathbf{s}_i\equiv \pi_i$, (\ref{eq:cost2}) reduces to (\ref{eq:cost1}) (ignoring the constant terms). Maximization of $F$ can now be achieved in an EM-like fashion. In the E-step we maximize $F$ w.r.t. $\mathbf{q}_i$, $\mathbf{s}_i$, while the M-step maximizes over the parameters $\Theta$ and $\pi$. We can see from (\ref{eq:cost2}) that the $F$ is maximized w.r.t $\mathbf{q}_i$ and $\mathbf{s}_i$ when the divergence terms vanish, therefore, $\mathbf{s}_i^\mathrm{opt} = \xi_{s_i} \pi_{i} \circ \pi_{N_i}$, $\mathbf{q}_i^\mathrm{opt} = \xi_{q_i} \mathbf{p}_{i} \circ \mathbf{p}_{N_i}$, where $\xi_{s_i}$ and $\xi_{q_i}$ are the normalization constants.

The M-step in not as straightforward, since direct optimization over $\Theta$ and $\pi$ is intractable and we resort to maximizing its lower bound. We define $\hat{\mathbf{s}}_i =(\mathbf{s}_i + \mathbf{s}_{N_i})$ and $\hat{\mathbf{q}}_i = (\mathbf{q}_i + \mathbf{q}_{N_i})$ and by Jensen's inequality lower-bound the divergence terms as
\begin{eqnarray}\label{eq:jensens}
     -D( \mathbf{s}_i \| \pi_i \circ \pi_{N_i}) \geq \hat{\mathbf{s}}_i^T \log \pi_i \nonumber\\ -D( \mathbf{q}_i \| \mathbf{p}_i \circ \mathbf{p}_{N_i}) \geq \hat{\mathbf{q}}_i^T \log \mathbf{p}_i,
\end{eqnarray}
where we have ignored the terms independent of $\pi_i$ and $\mathbf{p}_i$. Substituting (\ref{eq:jensens}) into (\ref{eq:cost2}) and collecting the relevant terms yields the following lower bound on the cost function (\ref{eq:cost2})
\begin{equation}\label{eq:cost3}  \small
    \hat F = \sum\limits_{i=1}^M [ {1 \over 2} (\mathbf{q}_i + \mathbf{q}_{N_i})^T \log( \mathbf{p}_i p(\Theta| \varphi_0) ) + {1 \over 2}(\hat{\mathbf{s}}_i + \hat{\mathbf{q}}_i)^T \log\pi_i ].
\end{equation}
Differentiating (\ref{eq:cost3}) w.r.t., $\pi_i$ and applying a Lagrange multiplier with the constraint $\sum\nolimits_{k} \pi_{ik}=1$, we see that $\hat F$ is maximized at $\pi_i^\mathrm{opt} = {1 \over 4} ( \hat{\mathbf{s}}_i + \hat{\mathbf{q}}_i )$.
Differentiating (\ref{eq:cost3}) w.r.t. the means and covariances of Gaussians, we obtain
\begin{eqnarray}
    \mu_k^\mathrm{opt} = \beta_k^{-1} [\Lambda_k(\sum\limits_{i=1}^M \hat{q}_{ik} \mathbf{y}_i^T) \Sigma_k^{-1} - \mu_{\mu_k}^T \Sigma_{\mu_k}^{-1} ]^T, \label{eq:MStep1}\\
    \Sigma_k^\mathrm{opt} =\beta_k^{-1} \sum\limits_{i=1}^M \hat{q}_{ik}(\mathbf{y}_i - \mu_k)(\mathbf{y}_i - \mu_k)^T, \label{eq:MStep2}
\end{eqnarray}
where we have defined $\beta_k=\sum\nolimits_{i=1}^M \hat{q}_{ik}$ and $\Lambda_k=(\Sigma_k^{-1} + \Sigma_{\mu_k}^{-1})^{-1}$.
An appealing property of the model (\ref{eq:cost2}) is that its E-step can be efficiently implemented through convolutions and Hadamard products. Recall that the calculation of the $i$-th pixel's neighborhood prior distribution $\pi_{N_i}$ entails a weighted combination of the neighboring pixel priors $\pi_{j}$. Let $\pi_{{\cdot k}}$ be the $k$-th component priors arranged in a matrix of image size. Then the neighborhood priors can be computed by the following convolution $\pi_{N_{\cdot k}}=\pi_{\cdot k}* \lambda$, where $\lambda$ is a discrete kernel with its central element set to zero and its elements summing to one. Let $\hat{\mathbf{s}}_{{\cdot k}}$, $\hat{\mathbf{q}}_{{\cdot k}}$ and $\mathbf{p}_{\cdot k}$ be the image-sized counterparts corresponding to sets of distributions $\{\hat{\mathbf{s}}_i \}_{i=1:M}$, $\{ \hat{\mathbf{q}}_i \}_{i=1:M}$ and $\{ {\mathbf{p}}_i \}_{i=1:M}$, respectively, and let $\lambda_1$ denote the kernel $\lambda$ in which the central element is set to one. Then the calculation of the $k$-th component priors $\pi^\mathrm{opt}_{\cdot k}$ for all pixels in the E-step can be written as
\begin{eqnarray}\label{eq:EstepPi_k}
    \hat{\mathbf{s}}_{{\cdot k}} = (\xi_{s \cdot} \circ \pi_{\cdot k} \circ (\pi_{\cdot k} * \lambda))*\lambda_1,\nonumber\\
    \hat{\mathbf{q}}_{{\cdot k}} = (\xi_{q \cdot} \circ \mathbf{p}_{\cdot k} \circ (\mathbf{p}_{\cdot k} * \lambda))*\lambda_1,\nonumber\\
    \pi^\mathrm{opt}_{\cdot k} = (\hat{\mathbf{s}}_{{\cdot k}} + \hat{\mathbf{q}}_{{\cdot k}})/4.
\end{eqnarray}
The EM procedure for fitting our generative model to the input image is summarized in Algorithm~\ref{alg:EM}.

\begin{algorithm}[h!]
\begin{algorithmic}[1]
 \REQUIRE {~}\\
    Pixel features $\mathbf{Y}=\{\mathbf{y}_i\}_{i=1:M}$, priors $\varphi_0$, initial values for $\Theta$ and $\pi$.
 \ENSURE {~}\\
    The estimated parameters $\pi^\mathrm{opt}$, $\Theta^\mathrm{opt}$ and the smoothed posterior $\{ \hat{\mathbf{q}}_{{\cdot k}} \}_{k=1:4}$.
 \\\hspace{-0.6cm}\textbf{Procedure:}
   \STATE Calculate the pixel posteriors $\mathbf{p}_{\cdot k}$ using the current estimates of $\pi$ and $\Theta$ for all $k$ (\ref{eq:measurementmodel}).
   \STATE Calculate the new pixel priors $\pi^\mathrm{opt}_{\cdot k}$ and posteriors $\hat{\mathbf{q}}_{{\cdot k}}$ for all $k$ using (\ref{eq:EstepPi_k}).
   \STATE Calculate the new parameter values $\Theta$ using (\ref{eq:MStep1}) and (\ref{eq:MStep2}).
   \STATE Iterate steps 1 to 3 until convergence.
\end{algorithmic}
\caption{\label{alg:EM}: The EM for semantic segmentation.}
\end{algorithm}

\section{Obstacle detection}\label{sec:obstacleDetectionAlgorithm}

We formulate the obstacle detection as a problem of estimating an image obstacle map, i.e., determining the pixels in the image that correspond to the sea while all the remaining pixels represent the potential obstacles. We therefore first fit our semantic model from Section~\ref{sec:segmentationModel} to the input image and estimate the smoothed a posteriori probability distribution $\hat{\mathbf{q}}_{{i k}}$ across the four semantic components for each pixel. An $i-\mathrm{th}$ pixel is classified as water if the corresponding posterior $\hat{\mathbf{q}}_{{i k}}$ reaches maximum for the water component among all four components. In our setting the component indexed by $k=1$ corresponds to water region, which results in the labeled image $B$ with the $i$-th pixel label $b_i$ defined as
\begin{equation}
 {b_i} = \left\{ {\begin{array}{*{20}{c}}
        1&;&\arg\max_k \hat{\mathbf{q}}_{{i k}} = 1\\
        0&;& otherwise
\end{array}} \right.  .
\end{equation}
Retaining only the largest connected region in the image $B$ results in the current obstacle image map $\hat{B}_t$. All blobs of non-water pixels within the connected water region are proclaimed as potential \textit{obstacles in the water}. This is followed by a nonmaxima suppression stage which merges detections that are located in close proximities (e.g., due to object fragmentation) to reduce multiple detections of the same obstacle. The water edge is extracted as the longest connected outer edge of the connected region corresponding to the water. The obstacle detection is summarized in Algorithm~\ref{alg:segdet} and visualized in Figure~\ref{fig:ourapproach}.

\begin{algorithm}[h!]
\begin{algorithmic}[1]
 \REQUIRE {~}\\
    Pixel features $\mathbf{Y}=\{\mathbf{y}_i\}_{i=1:M}$, priors $\varphi_0$, estimated model from previous time-step $\Theta_{t-1}$ and $\hat{\mathbf{q}}_{t-1}$.
 \ENSURE {~}\\
    Obstacle image map $\hat{B}_t$, water edge $\mathbf{e}_t$, detected objects $\{ \mathbf{o}_i \}_{i=1:N_\mathrm{obj}}$, model parameters $\Theta_{t}$ and $\hat{\mathbf{q}}_{t}$.
 \\\hspace{-0.6cm}\textbf{Procedure:}
    \STATE Initialize the parameters of $\Theta_{t}$ and $\mathbf{\pi}_{t}$ according to Section~\ref{sec:parInit}.
    \STATE Apply the Algorithm~\ref{alg:EM} to fit the model $\Theta_{t}$ and $\hat{\mathbf{q}}_{t}$ to the input data $\mathbf{Y}$.
    \STATE Calculate the new obstacle image map $\hat{B}_t$ and for interpretation also the water edge $\mathbf{e}_t$ and the obstacles in water $\{ \mathbf{o}_i \}_{i=1:N_{obj}}$.
\end{algorithmic}
\caption{\label{alg:segdet}: The obstacle image map estimation and obstacle detection algorithm.}
\end{algorithm}

\subsection{Initialization}\label{sec:parInit}

\begin{figure}
        \centering
            \includegraphics[width=\halfpagewidth]{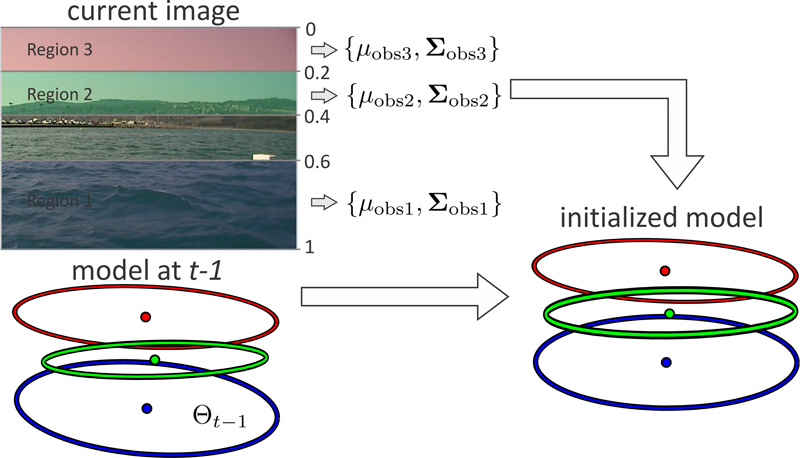}
        \caption{\label{fig:imagePart} Illustration of image partitioning for extraction of $\Theta_\mathrm{obs}$ components and combination
        with the model from the previous time-step $\Theta_{t-1}$ for initialization of the EM.}
\end{figure}

The Algorithm~\ref{alg:EM} requires initial values for the parameters $\Theta_t$ and $\pi_t$. At the first frame, when no other prior knowledge exists, we construct the initial distribution by vertically splitting the image into three regions $\{0,0.2\}$, $\{0.2,0.4\}$ and $\{0.6,1\}$, written in proportions of the image height (see Figure~\ref{fig:imagePart}). A Gaussian is computed from each region, thus forming the \textit{observed} components $\Theta_\mathrm{obs}=\{ \mu_{\mathrm{obs}k}, \mathbf{\Sigma}_{\mathrm{obs}k}, w_{\mathrm{obs}k}\}_{k=1:3}$. The prior over all pixels is initialized to equal probabilities for the three components, while the prior on the uniform component is set to a low constant value (see Section~\ref{sec:implementDetails}). These parameters are used to initialize the EM in the Algorithm~\ref{alg:EM}.

The shape for the vertical splits in Figure~\ref{fig:imagePart} should ideally follow the position (and inclination) of true horizon for optimal initialization of the parameters. An estimate of the true horizon depends on the camera placement and can ideally be obtained externally from an IMU sensor, but the per-frame IMU measurements are not available in the dataset that is used in our evaluation (Section~\ref{sec:modddataset}). Therefore, an assumption is made that the horizon, as well as edge of water, is usually located within the region $\{0.4,0.6\}$ of image height, which is the reason this region is excluded from computation of parameter initial values. Making no further assumptions regarding the proportion between components in the final segmentation, equal regions (2 and 3) are used to initialize the parameters of the component 2 and 3. The assumption on region splitting is often violated in our dataset from Section~\ref{sec:modddataset} due to boat inclination at turning maneuvers, due to boat tilting forward and backward, and since the camera might not have been mounted to exactly the same spot in assembling the boat after transportation to the test site during the several months that the dataset was taken. Nevertheless, the segmentation algorithm is robust enough to handle the non-ideal initializations as long as there are no extreme deviations, like the boat toppling or riding on extremely high waves (the small coastal USVs are actually not even designed to physically endure these extreme weather conditions).

During the USV's operation, we can exploit the continuity of sequential images in the videostream by using the parameter values of the converged model from the previous time-step for initialization of the EM algorithm in the current time-step. To reduce possible propagation of errors stemming from false segmentations in the previous time-steps, a zero-order soft reset is applied in the initialization of the EM in each time-step. In particular, the EM is initialized by merging the $\Theta_\mathrm{obs}$ with $\Theta_{t-1}$. The parameters of the $k$-th component, $\{ \mu_{\mathrm{init}k}, \Sigma_{\mathrm{init}k} \}$, are initialized by forming a weighted two-component mixture model from the $k$-th components in $\Theta_\mathrm{obs}$ and $\Theta_{t-1}$, and approximating them by a single component by matching to first two moments of the distributions (see, e.g., Kristan et al.~\cite{KristanPR11,Kristan2013a,Wang2014}). The weights $\alpha$ and $1-\alpha$ for $\Theta_{t-1}$ and $\Theta_\mathrm{obs}$, respectively, can be used to balance the contribution of each component. The priors $\pi_t$ over the pixels are initialized by the smoothed posterior $\hat{\mathbf{q}}_{{\cdot k}}$ from the previous time-step. The initialization is sketched in Figure~\ref{fig:imagePart}.

\section{Implementation details}\label{sec:implementDetails}

In our application, the measurement at each pixel is encoded by a five-dimensional feature vector
$\mathbf{y}_i = [ i_x, i_y, i_{c1},i_{c2},i_{c3}]$, where $(i_x, i_y)$ are the $i$-th pixel coordinates and the $(i_{c1},i_{c2},i_{c3})$ are the pixel's color channels. We have determined in a preliminary study that we achieve sufficiently good obstacle detection by first performing detection on a reduced-size image of $50 \times 50$ pixels and then rescaling the results to the original image size. The rescaling was set to match the lower scale of objects of interest, as smaller objects do not present danger to the USV. Such approach drastically speeds up the algorithm to approximately {10ms per frame} in our experiments. The uniform distribution component in~(\ref{eq:measurementmodel}) is defined over the image pixels domain and returns equal probability for each pixel. Assuming that all color channels are constrained to the interval $[0,1]$, the value of the uniform distribution is $\mathcal{U}(\mathbf{y}_i)={1 \over 50^2}$ at each pixel for our rescaled image. The EM optimization requires specification of the convolution kernel $\lambda$. Note that the only constraint on the convolution kernel is that its central element is set to zero and all elements sum to one. We use a Gaussian kernel with its central element set to zero and set the size of the kernel to $2\%$ of image size, which results in a $3 \times 3$ pixels kernel. Recall from Section~\ref{sec:parInit} that the parameter $\alpha$ influences the soft-reset of the parameters used to initialize the EM. In our implementation, a slightly larger weight is given to the parameters estimated at the previous time-step by setting $\alpha=0.6$.

\subsection{Learning the weak priors}

The spatial components in the feature vector play a dual role. First, they enforce to some extent the spatial smoothness of the segmentation on their own. Second, they lend means to weakly constraining the Gaussian components such that they reflect the three dominant semantic image parts. This is achieved by the weak priors $p(\Theta | \varphi_0)=\prod\nolimits_{k=1}^3 \phi(\mu_k | \mu_{\mu_k}, \Sigma_{\mu_k})$ on the Gaussian means. Since the locations and shape of semantic components vary significantly with the views, we indeed select weak priors, which are estimated using the training set from our database (see Section~\ref{sec:modddataset}). Given a set of training images, the prior of the $k$-th component is estimated by extracting the features, i.e. sets of $\mathbf{y}_i$, corresponding to the $k$-th component from all images and fit a single Gaussian to them.
Note that, in general, there is a chance that the training images might bias the horizontal location of the estimated Gaussian to the left or right part of the image. In this case, we could constrain the horizontal position of the Gaussians to be in the middle of the image, however, we have observed that the components of the prior estimated from our dataset are sufficiently centered and we do not apply any such constraints.

\begin{figure}[h]
        \centering
            \includegraphics[width=\halfpagewidth]{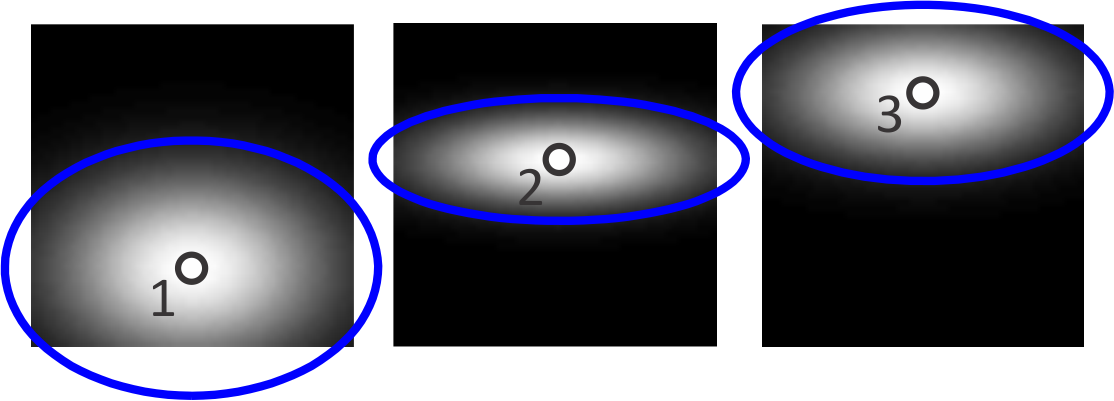}
        \caption{\label{fig:priorvis} Visualization of the spatial part of the Gaussians in the weak priors over our three semantic components. From left to right: bottom-most, middle and top-most component prior.}
\end{figure}

Examples of the spatial parts of the priors estimated from the training set of the dataset presented in Section~\ref{sec:modddataset} are shown in Figure~\ref{fig:priorvis}. Our algorithm, as well as the learning routine, was implemented in Matlab -- a reference code is publicly available at~\footnote{\url{http://www.vicos.si/Research/UnmannedSurfaceVehicles}}.

\section{Marine obstacle detection dataset}\label{sec:modddataset}

With lack of sufficiently large publicly available annotated dataset to test our method, we have constructed our own dataset, which we call the Marine obstacle detection dataset (Modd). The Modd consists of 12 video sequences, providing in total 4454 fully annotated frames with resolution of 640 x 480 pixels. The dataset is made publicly available along with the annotations and Matlab evaluation routines from the MODD homepage\footnote{\url{http://www.vicos.si/Downloads/MODD}}.

The video sequences have been recorded from multiple platforms, most of them from the small 2.2 meter USV\footnote{A video of our USV is available online from the MODD homepage.}
(see Figure~\ref{fig:ourapproach}). The USV was developed by Harpha Sea, d.o.o. Koper, and is based on catamaran hull design and powered by electrical, LiPo battery powered, steerable thrust propeller. It can reach the maximum speed of 2.5 m/s and has extremely small turn radius. Steering and route planning are handled by ARM-powered MCU with redundant power supply. For navigation, the MCU relies on microelectromechanical inertial navigation unit (MEMS IMU), solid-state digital compass and differential GPS. USV has two different communication channels to the shore (high- and low-bandwith) and its mission can be programmed remotely. An Axis 207W camera was placed on the USV approximately 0.7 m above the water surface, looking in front of the vessel, with an approximately 55$^{\circ}$ field of view. Camera has been set up to automatically adjust to the variations in lighting conditions. Since the boat was being reassembled between the runs over several months, the placement of the camera varies slightly across the dataset.

\begin{figure}
        \centering
            \includegraphics[width=8.5cm]{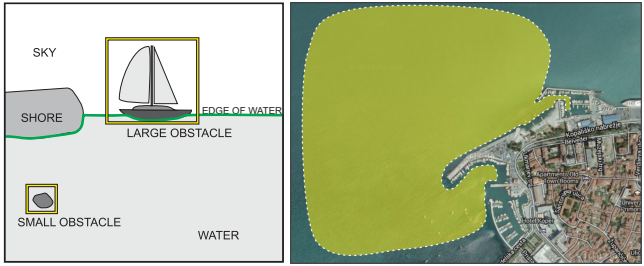}
        \caption{\label{fig:waterobjects}Left: Scene representation used for evaluation. Right: The dashed outline denotes the region in the coastal waters of Koper, Slovenia (gulf of Trieste), where the dataset was captured.}
\end{figure}

The video sequences have been acquired in the gulf of Trieste, specifically in the port of Koper, Slovenia, (Figure~\ref{fig:waterobjects}) over a period of months at different times of day under different weather conditions. The USV was manually operated by a human pilot and effort was made to simulate realistic navigation, including threats of collision. The pilot was instructed to deliberately drive in a way to simulate situations in which an obstacle might present a danger to the USV. This includes obstacles being very close to the boat as well as situations in which the boat was heading straight towards an obstacle for a number of frames.

\begin{figure}[htb]
        \centering
            \includegraphics[width=8.5cm]{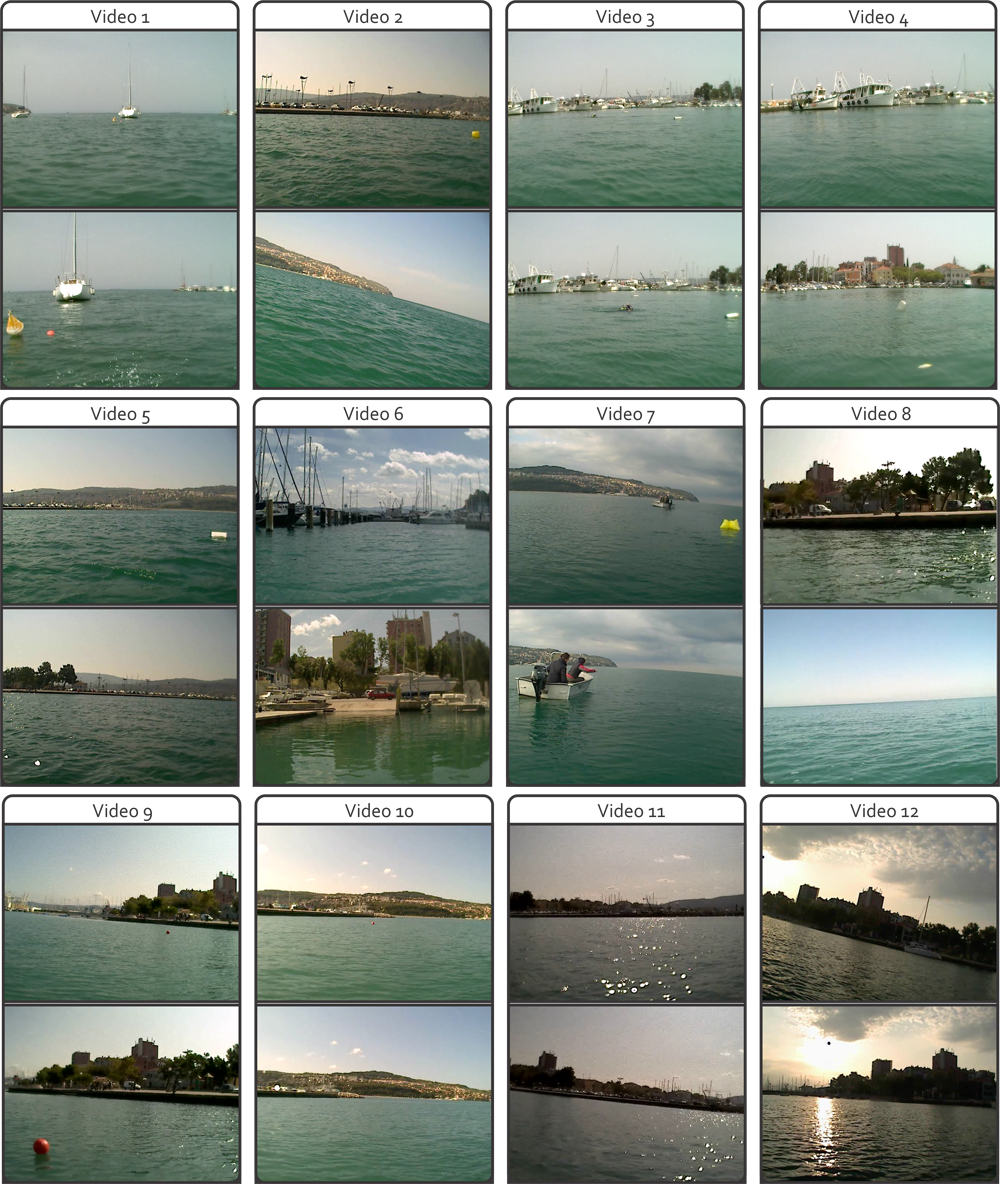}
        \caption{\label{fig:Modd_pictures} Examples of images taken from the videos in the Modd. The first ten videos are for normal conditions, while the last two depict extreme conditions. For each video we show two images for better impression of the video content.}
\end{figure}

The first ten videos in the dataset are meant for evaluation of the obstacle-map estimation algorithms under normal conditions. These videos still vary quite significantly between each other and simulate conditions under which the USV is expected to operate. We thus term these ten videos as \textit{normal conditions} and we show some examples of these videos in the first ten images from Figure~\ref{fig:Modd_pictures}. The last two videos were meant for analysis in situations in which the boat is directly facing the sun. This causes extreme changes in the automatic shutter and camera setting, resulting in significant changes of contrast and color of all three semantic components. Facing the sun also generates significant amount of fragmented glitter, while sometimes it shows up as a larger, fully connected region of the reflected sun. We thus denote these last two videos as \textit{extreme conditions}. Some examples are shown in the last two images of Figure~\ref{fig:Modd_pictures}.

Each frame is annotated manually by a polygon denoting the edge of water and bounding boxes are placed on \emph{large obstacles} (those that straddle the water edge) and \emph{small obstacles} (those that are fully surrounded by water). See Figure~\ref{fig:waterobjects} for illustration. The annotation was made by a human annotator and all annotations on all images of the dataset were later verified by an expert. To allow a fast overview of the annotations by the potential users of the dataset, the dataset provides a rendered video with annotations overlay, for each test sequence in the dataset -- these videos are included as part of the dataset and available from the dataset homepage as well.

\begin{figure}
        \centering
            \includegraphics[width=7.5cm]{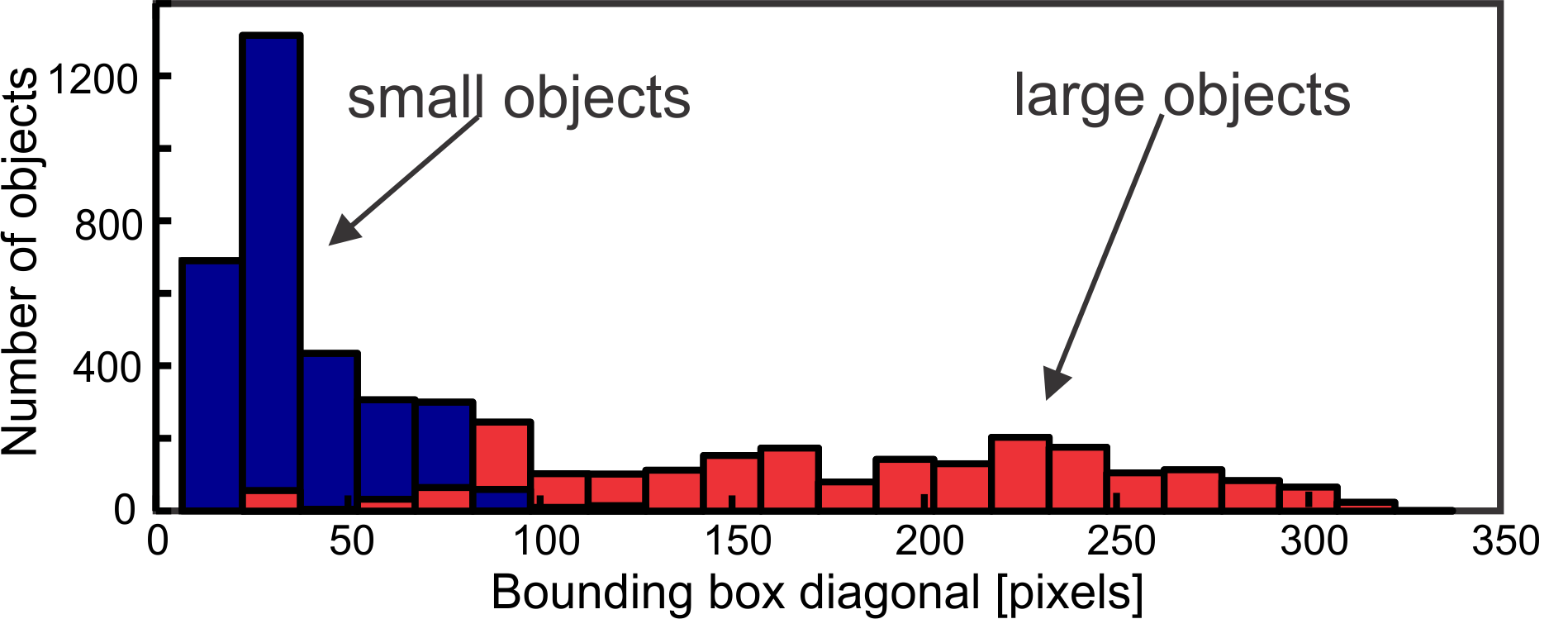}
        \caption{\label{fig:pdfObstacleSize}Distributions of the floating obstacle sizes labelled as large and small shown in red and blue, respectively.}
\end{figure}

\begin{figure}
        \centering
            \includegraphics[width=8cm]{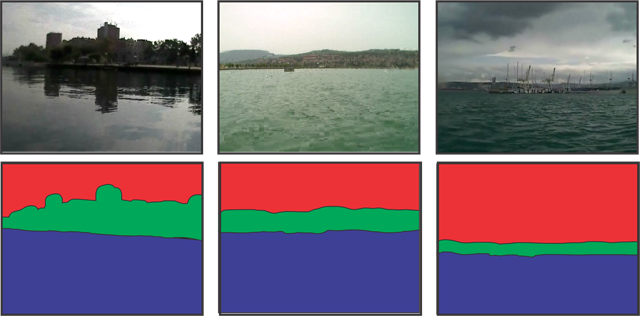}
        \caption{\label{fig:manualAnnot}Examples of training images along with their manual label masks. The blue, green and red color correspond to the labels for bottom, middle and the top semantic component, respectively.}
\end{figure}

In the following some general statistics of the dataset are provided. The boat was driving within 200 meters from the shore, and most of the depicted obstacles are in this zone. Out of 12 image sequences in the dataset, nine contain either large or small obstacles, one contains only annotated sea edge, and two contain glitter annotations, sea edge annotations, and no objects. The number of objects per frame is exponentially distributed with the average 1.1 and variance 1.23. The distribution of the annotated size of small and large obstacles is shown in Figure~\ref{fig:pdfObstacleSize}. For the algorithms that require training or validation of their parameters, we have compiled a collection of twenty images in which we manually annotated the pixels corresponding to three semantic components. Figure~\ref{fig:manualAnnot} shows some examples of images and the corresponding annotations.

\subsection{The evaluation protocol}

The evaluation protocol is designed to reflect the two distinct challenges that the USVs face: the water edge (shoreline/horizon) detection and obstacle detection. The former is measured as the root mean square error (RMSE) of the water edge position ($Edg$), and the latter is measured via the efficiency of \emph{small object} detection, expressed as precision ($Prec$), recall ($Rec$), F-score ($F$) and the average number of false positives per frame (${aFP}$).

To evaluate RMSE in water edge position, ground truth annotations were used in the following way. A polygon, denoting the water surface was generated from water edge annotations. Areas, where \emph{large obstacles} intersect the polygon, were removed. Note that, given the scene representation, shown in Figure~\ref{fig:waterobjects}, one cannot distinguish between large obstacles (e.g., large ships) and stationary elements of the shore (e.g., small piers). This way, a refined water edge was generated. For each pixel column in the full-sized image, a distance between water edge, as given by the ground truth and as determined by the algorithm, was calculated. These values are summarized into a single $Edg$ value by averaging across all frames and videos.

\begin{figure}
        \centering
            \includegraphics[width=7.5cm]{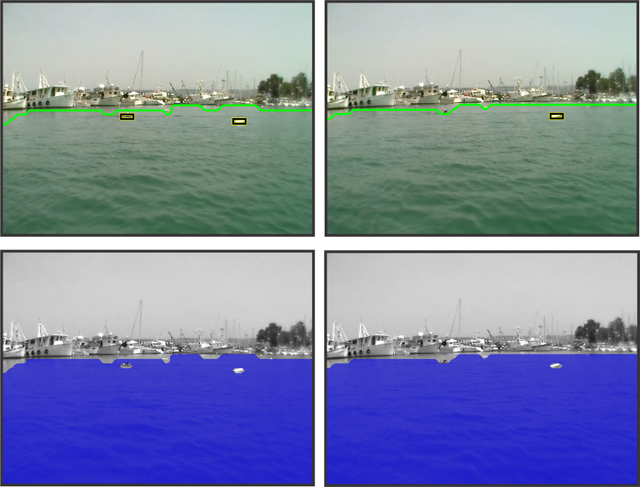}
        \caption{\label{fig:edgeProblems} Images with a scubadiver emerging from the water just at the observed water edge -- the upper row shows the water edge and detected obstacles in the water while the bottom row shows the water mask. In the left-hand images, the estimated water edge runs above the diver and the scubadiver is explicitly detected. In the right-hand images the edge runs below the scubadiver, which prevents explicit detection, eventhough the region corresponding to the scubadiver is in fact detected as a part of the obstacle.}
\end{figure}

The evaluation of object detection follows the recommendations from the PASCAL VOC challenges by Everingham et al.~\cite{Everingham10}, with small, application-specific modification: all small obstacles (provided as a ground truth or detected) that are closer to the annotated water line than 5\% of the image height, are discarded prior to evaluation on each frame. This aims to address situations where a detection may oscillate between fully water-enclosed obstacle, and the ``dent'' in the shoreline, resulting in false negatives. Figure~\ref{fig:edgeProblems} shows an example with two images of a scubadiver emerging from the water. Note that in both images, the segmentation successfully labeled the scubadiver as an obstacle. But in the left-hand image we obtain an explicit detection, since the estimated water edge runs above the scubadiver. In the right-hand image the edge runs below the scubadiver and we do not get explicit detection, eventhough the algorithm successfully labeled the scubadiver's region as being an obstacle. Note that the proposed treatment of near-edge detections/ground-truths is also consistent with the problem of obstacle avoidance -- the USV is concerned primarily with the avoidance of the obstacles in its immediate vicinity. In counting false positives (FP), true positives (TP) and false negatives (FN), we follow the methodology of PASCAL VOC, with the minimum overlap set to 0.3. FP, TP and FN were used to calculate precision ($Prec$), recall ($Rec$), F-score ($F$) and average false positives per frame (${aFP}$).

\section{Experiments}\label{sec:experiments}

In the following we will denote our obstacle image-map estimation method (Algorithm~\ref{alg:segdet}) as the semantic-segmentation model ($\mathrm{SSM}$). The experimental analysis was split into three parts. In the first part we evaluate the influence of the different color spaces on the $\mathrm{SSM}$'s performance. In the second and third part we analyze how various elements of $\mathrm{SSM}$ affect its performance and compare it to the alternative methods. All experiments were performed on a desktop PC with 3.06 GHz Intel Xeon E5-1620 CPU in a single thread in Matlab.

\subsection{Influence of the color space}

The aim of the first experiment was to evaluate how the different colorspaces affect the segmentation performance. We have therefore performed experiments in which the feature vector $\mathbf{y}_i$ (Section~\ref{sec:implementDetails}) was calculated from RGB, HSV, Lab and YCrCb colorspace. For each of the selected colorspaces, the weak priors were learned from the training images on the Modd dataset (Section~\ref{sec:modddataset}). All experiments were performed on all twelve testing videos from the Modd dataset. The results are shown in Table~\ref{tbl:resultsColor}.

\begin{table}\caption{\label{tbl:resultsColor}Effects of the colorspace on the SSM segmentation performance using all twelve videosequences from the Modd. The results are given by reporting the average performance with a standard deviation in brackets: Edge of water estimation error, precision, recall, F measure and average false positives, $\mathrm{Edg}$, $\mathrm{Prec}$, $\mathrm{Rec}$, $F$, ${aFP}$. For each performance measure, the best performing method is marked in bold.}\centerline{
\begin{tabular}{|@{\extracolsep{\fill}}c |c c c c c|}
\hline
 colorspace & $\mathrm{Edg} [\mathrm{pix}]$ & $\mathrm{Prec}$ & $\mathrm{Rec}$ & $F$  & ${aFP}$ \\
\hline
 RGB   & 10.7(5.8)         & 0.874          & 0.756          & 0.806          & {0.039}        \\
 HSV   & 12.7(8.3)         & 0.821          & 0.688          & 0.738          & 0.041         \\
 Lab   & \textbf{9.3}(5.1) & {0.878}        & {0.768}        & {0.815}        & {0.039}       \\
 YCrCb & {9.5}(5.5)        & \textbf{0.885} & \textbf{0.772} & \textbf{0.819} & \textbf{0.039} \\
 \hline
\end{tabular}}
\end{table}

The results show that best performance is achieved with the YCrCb and Lab colorspace, which is not surprising, since these colorspaces are known to better cluster visually-similar colors. Similar is true for the HSV space, but that space suffers from the circular property of the Hue component (i.e., red color is on the left-most and right-most part of the Hue spectrum). With respect to the edge of the water estimation, the lowest error is achieved when using the Lab colorspace, while only a slightly worse performance is obtained with the YCrCb colorspace. On all other measures, the YCrCb colorspace yields best results, although comparable to the Lab colorspace. While the results are worse when using the RGB or the HSV colorspace, we note that these results do not exhibit drastically poorer performance, which speaks of a level of robustness of the SSM to the choice of the colorspace. Nevertheless, given the results in Table~\ref{tbl:resultsColor}, we select the YCrCb and use this colorspace in the subsequent experiments.

\subsection{Comparison to alternative approaches}

Given a fixed colorspace, we are left with evaluation of how much each part of our model contributes to the final performance. We have therefore also implemented two variants of our approach, which we denote by $\mathrm{UGM}$ and $\mathrm{UGM_{col}}$. In contrast to $\mathrm{SSM}$, the $\mathrm{UGM}$ and $\mathrm{UGM_{col}}$ do not use the MRF constraints and are in this respect only mixtures of three Gaussians with priors on their means and with a uniform component. A further difference between $\mathrm{UGM}$ and $\mathrm{UGM_{col}}$ was that $\mathrm{UGM_{col}}$ ignored the spatial information in visual features and relied only on color.

Note that the $\mathrm{SSM}$ is conceptually similar to the Grab-cut algorithm from Rother et al.~\cite{Rother2004} for binary segmentation, but with distinct differences. In the Grab-cut, the user provides a bounding box roughly containing the object, thus initializing the segmentation mask. Two visual models using a GMM are constructed from this segmentation mask. One for the object and one for the background. A MRF is then constructed over the pixel grid and graph cut from Boykov et al.~\cite{Boykov2001} is used to infer an improved segmentation mask. This procedure is then iterated until convergence. There are significant differences between the proposed $\mathrm{SSM}$ and the Grab-cut from~\cite{Rother2004}. In contrast to the user-provided bounding box in~\cite{Rother2004}, the $\mathrm{SSM}$'s weak supervision comes from the initialization of the parameters from the previous time-step and from the weak priors. The second distinction is that our approach does require explicit estimation of the segmentation mask to refine the mixture model. This allows for a better propagation of uncertainty during the iteration of the algorithm, leading to improved segmentation.

To further evaluate contributions of the particular MRF optimization of our SSM, we have implemented a variant of the Grab-cut algorithm, which uses our semantic mixture model, but applies graph-cuts for optimization over the MRF. The resulting obstacle-map estimation tightly follows Algorithm~\ref{alg:EM} and Algorithm~\ref{alg:segdet} with a slight modification of the Algorithm~\ref{alg:EM}: After each epoch of the EM, we apply the graph-cut from Bagon~\cite{BagonMatlabGraphcut} to segment the image into a water/non-water mask. This mask is then used as in the original Grab-cut to refine the mixture model. We use exactly the same weakly-constrained mixture model as in $\mathrm{SSM}$, and the YCrCb colorspace for fair comparison, and call this approach the Grab-cut model $\mathrm{GCM}$.

We have compared our approach also to the general segmentation approaches, namely the superpixel-based approach from Li et al.~\cite{Li2012}, $\mathrm{SPX}$, and a graph-based segmentation algorithm from Felzenswalb and Huttenlocher~\cite{Felzenszwalb2004}, $\mathrm{FZH}$.

For fair comparison, all the algorithms were executed on the $50\times50$ images. We have experimented with the parameters of $\mathrm{GCM}$ and $\mathrm{FZH}$ and have set them to optimal performance for our dataset. Since $\mathrm{FZH}$ was designed to run on larger images, we have also performed the experiments for $\mathrm{FZH}$ on full-sized images -- we denote this variant by $\mathrm{FZH_{full}}$. We have performed the comparative analysis separately for the \textit{normal} and \textit{extreme} conditions.

\subsubsection{Performance under normal conditions}

\begin{figure}
        \centering
            \includegraphics[width=8.5cm]{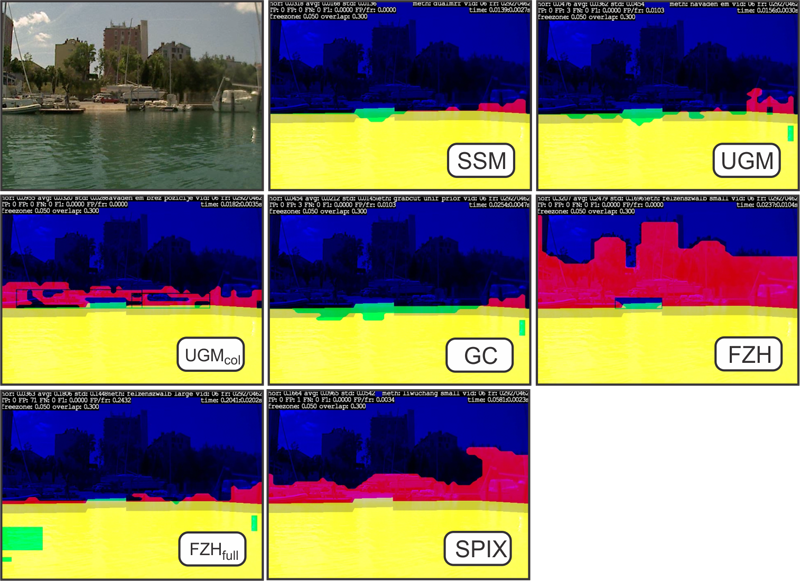}
        \caption{\label{fig:visualComparison1} Qualitative comparison of the different obstacle-map estimation approaches. The upper left-most image is the original input image, followed by the results for $\mathrm{SSM}$, $\mathrm{UGM}$, $\mathrm{UGM_{col}}$, $\mathrm{GCM}$, $\mathrm{FZH_{}}$, $\mathrm{FZH_{full}}$ and  $\mathrm{SPX_{}}$.This image is best viewed in color. Please see text for the description of the color codes.}
\end{figure}

The results of the experiments on the \textit{normal conditions} part of the Modd are summarized in Table~\ref{tbl:resultsNormal}, while Figure~\ref{fig:visualComparison1} shows an example of  typical segmentation masks from the compared algorithms. The segmentation results in these images are color coded as follows. The original image is represented only by the blue channel, manual water annotations are shown in the green channel, and algorithm-generated water segmentation is shown in the red channel. Therefore, the cyan region shows the area, which has been annotated as water, but has not been segmented as such by the algorithm (bad). The magenta region shows the area, which has not been annotated as water, but has been segmented as such by the algorithm (bad). The yellow area shows the area which has been annotated as water and has been segmented as such by the algorithm (good), and blue region shows the area which has not been annotated as water and has not been segmented as such (good). Finally, the darker band under the annotated edge of the water in all colors shows the \textit{ignore} region, where evaluation of small obstacle detection does not take place.

\begin{table}[htb]\caption{\label{tbl:resultsNormal} Comparison of various methods under \textit{normal conditions}. The results are given by reporting average performance with a standard deviation in brackets: Edge of water estimation error in pixels ($\mathrm{Edg}$), precision ($\mathrm{Prec}$), recall ($\mathrm{Rec}$), F measure ($F$), average false positives ($aFP$) and time in [ms] ($t$).}\centerline{
\begin{tabular}{@{\extracolsep{\fill}}l |c c c c c c}
 & $\mathrm{Edg}$ & $\mathrm{Prec}$ & $\mathrm{Rec}$ & $F$  & {aFP} & $t$\\
\hline
 $\mathrm{SSM}$ & \textbf{9.2}(4.9) & \textbf{0.885} & \textbf{0.772} & \textbf{0.819} & \textbf{0.039} & \textbf{10}(0) \\
 $\mathrm{GCM}$ & 10.9(5.6) & 0.718 & 0.686 & 0.695 & 0.121 & 17(3) \\
 $\mathrm{UGM}$ & 10.5(6.1) & 0.742 & 0.706 & 0.717 & 0.109 & 11(2) \\
 $\mathrm{UGM_{col}}$ & 16.4(9.0) & 0.614 & 0.504 & 0.549 & 0.122 & 11(3) \\
 $\mathrm{FZH}$ & 90.0(65.7) & 0.727 & 0.523 & 0.551 & 0.053 & 16(1) \\
 $\mathrm{FZH_{full}}$ & 34.2(41.4) & 0.410 & 0.747 & 0.488 & 0.697 & 199(3) \\
 $\mathrm{SPX}$ & 66.4(34.7) & 0.007 & 0.001 & 0.001 & 0.090 & 54(1) \\
 \end{tabular}}
\end{table}

Recall that in contrast to the $\mathrm{SSM}$, the $\mathrm{UGM}$ and $\mathrm{UGM_{col}}$ do not impose a Markov random field constraint. In Figure~\ref{fig:visualComparison1} this clearly leads to a poorer segmentation, resulting in false positive detections of obstacles as well as significant over-segmentations. Quantitatively, the poorer performance is reflected in Table~\ref{tbl:resultsNormal} as a lower $F$-measure, higher average number of false positives and larger edge of the water estimation error. Compared to $\mathrm{SSM}$, we observe a significant drop in detection quality of the $\mathrm{UGM}$, especially precision. This speaks of importance of the local labeling constraints imposed by the MRF in the $\mathrm{SSM}$. The performance further drops with $\mathrm{UGM_{col}}$, which implies that spatial components in the feature vectors bear important information for proper segmentation as well. On the other hand, the $\mathrm{GCM}$ does impose a MRF, however, the segmentation is still poorer than with the $\mathrm{SSM}$. We believe that the main reason for this is that the $\mathrm{GCM}$ applies graph-cuts to perform hard segmentation during EM updates. On the other hand, the $\mathrm{SSM}$ optimizes the cost function within a single EM framework, thus avoiding the need for hard segmentations during the EM steps, which leads to a better final result. By far the worst segmentation results are obtained by the $\mathrm{FZH}$, $\mathrm{FZH_{full}}$ and $\mathrm{SPX}$ segmentation methods. Note that while these segmentation methods do assume some local consistency of segmentation, they still perform poorer than the $\mathrm{SSM}$. The improved performance of $\mathrm{SSM}$ can be attributed exclusively to our formulation of the segmentation model within the graphical model from Figure~\ref{fig:graphicalmodel}.

\begin{figure*}
        \centering
            \includegraphics[width=16cm]{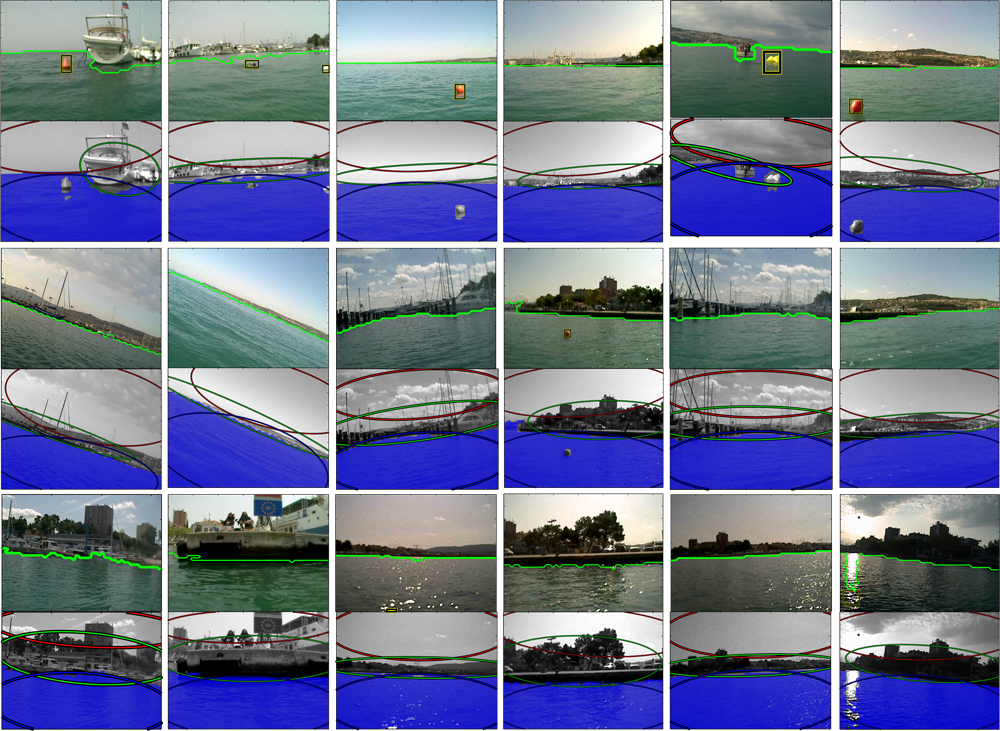}
        \caption{\label{fig:goodsegmentation} Qualitative examples of water segmentation and obstacle detection. We show for each image the detected edge of the sea in green and the detected obstacle by a yellow rectangle. Below each image we also show the spatial part of the three semantic components as three Gaussian ellipses and the portion of the image segmented as water in blue. This figure is best viewed in color.}
\end{figure*}

Figure~\ref{fig:goodsegmentation} shows further examples of SSM segmentation maps (the first fourteen images), the spatial part of the Gaussian mixture and the detected objects in water. The appearance and texture of the water varies significantly between the various scenes, and the same is true for the other two semantic components. The images also vary in the scene composition in that the vertical position as well as the attitude of the water edge (see second row in Figure~\ref{fig:goodsegmentation}) vary significantly. Nevertheless, note that the model is able to adapt well to these compositions and successfully decomposes the scene into water regions, in-water obstacles and fairly well delineates the water edge.

Our algorithm performed (segmentation+detection) at a rate higher than 70 frames per second. Most of the processing was spent on fitting our semantic model and obstacle-map estimation ($10$ ms), while $4$ ms was spent on the obstacle detection. For fair comparison of segmentation algorithms, we report in the Table~\ref{tbl:resultsNormal} only the times required for the obstacle-map estimation. Although note that the obstacle detection part did require more processing time for the methods that delivered poor segmentation masks with more false positives. On average, our EM algorithm in $\mathrm{SSM}$ converged in approximately three iterations. Note that the graph cut routine in $\mathrm{GCM}$, part of $\mathrm{SPX}$ and the $\mathrm{FZH}$ were implemented in C and interfaced to Matlab, while all the other variants were entirely implemented in Matlab. Therefore, the computational time results for segmentations are not directly comparable among the methods, but still offer a level of insight. In terms of processing time, the $\mathrm{SSM}$'s segmentation was the fastest, running at 100 frames per second. The $\mathrm{UGM_{col}}$ and $\mathrm{UGM}$ performed approximately as fast as $\mathrm{SSM}$, followed by $\mathrm{GCM}$, $\mathrm{FZH_{}}$, $\mathrm{SPX}$ and $\mathrm{FZH_{full}}$. We conclude that the $\mathrm{SSM}$ came out on top as the fastest method that also achieved the best detection performance as well as accuracy.

\subsubsection{Performance under extreme conditions}

We were interested in measuring two properties of the algorithms under conditions when the boat is facing the sun. In particular, were interested in measuring how the sun affects the edge-of-water estimation and how the glitter affects the detection. We have therefore repeated two variants of the experiments on the videos in Modd denoted by the \textit{extreme conditions} (videos 11 and 12 in Figure~\ref{fig:Modd_pictures}). In the first variant, we ignored any detections in the regions that were denoted as glitter regions in ground truth. In the second variant, all detections were accounted for. Note that the videos denoted as \textit{extreme conditions} do not contain any objects, therefore there were no true positives and any detected object was a false positive. Because of this, we present in the results (Table~\ref{tbl:resultsExtreme}) only the edge of water estimation error and the average number of false positives (by definition, both, accuracy and precision, would be zero in such case).

In terms of edge of water estimation, the $\mathrm{UGM_{col}}$ slightly outperforms the $\mathrm{SSM}$. The $\mathrm{UGM_{col}}$ ignores the spatial information and generally oversegments the regions close to the shoreline (as seen in Figure~\ref{fig:visualComparison1}), which in this case actually reduces the error compared to $\mathrm{SSM}$. The reason is that the $\mathrm{SSM}$ attributes the upper part of the sun reflection at the shoreline in video 11 (Figure~\ref{fig:Modd_pictures}) to an obstacle instead of the water. When ignoring the glitter region, the $\mathrm{SSM}$ outperforms the competing methods by not detecting any false positives (zero ${aFP}_\mathrm{ignore}$), while the competing methods exhibit larger values of the false positives. When considering also the glitter region, the number of false positives only slightly increases for the $\mathrm{SSM}$, while this increase is considerable for the other methods. Note that in this case the $\mathrm{SSM}$ again significantly outperforms the other methods, except for $\mathrm{SPX}$. The reason is that the $\mathrm{SPX}$ actually fails by grossly oversegmenting the water region, thus assigning almost all glitter to that region. However, looking at the results of the edge estimation, we can also see that this oversegmentation actually consumes also a part of the shoreline, thus leading to poor overall segmentation. Among the remaining methods, the $\mathrm{SSM}$ again achieves the lowest average false positive rate. Given these results we conclude that the $\mathrm{SSM}$ is much more robust to extreme conditions than the competing methods, while still offering good segmentation results. Some examples of segmentation with $\mathrm{SSM}$ are shown in the last four images of Figure~\ref{fig:goodsegmentation}. Even in these harsh conditions the model is able to interpret the scene well enough with few false obstacle detections. For more illustrative examples of our method and segmentations, please consult the additional online material at \url{http://box.vicos.si/matejk/smc/index.htm}.

\begin{table}\caption{\label{tbl:resultsExtreme} Comparison of various methods under \textit{extreme conditions}. We show the results for edge of water estimation error $\mathrm{Edg}$ and average false positives when ignoring glitter and when counting the glitter as false positives, ${aFP}_\mathrm{ignore}$ and ${aFP}_\mathrm{account}$, respectively.}\centerline{
\begin{tabular}{|@{\extracolsep{\fill}}l |c c c|}
\hline
 & $\mathrm{Edg} [\mathrm{pix}]$ &  ${aFP}_\mathrm{ignore}$ & ${aFP}_\mathrm{account}$ \\
\hline
 $\mathrm{SSM}$       & 11.3(10.0) & \textbf{0.000} & 0.134 \\
 $\mathrm{GCM}$        & 16.1(15.3) & 0.010 & 0.919 \\
 $\mathrm{UGM}$       & 12.4(11.3) & 0.007 & 0.932 \\
 $\mathrm{UGM_{col}}$ & \textbf{8.1}(5.1)   & 0.019 & 0.308 \\
 $\mathrm{FZH}$       & 65.3(48.7) & 0.003 & 0.233 \\
 $\mathrm{FZH_{full}}$& 46.4(51.6) & 0.159 & 2.056 \\
 $\mathrm{SPX}$       & 49.2(48.6) & 0.015 & \textbf{0.019} \\
\hline
 \end{tabular}}
\end{table}

\subsubsection{Failure cases}

\begin{figure}
        \centering
            \includegraphics[width=8cm]{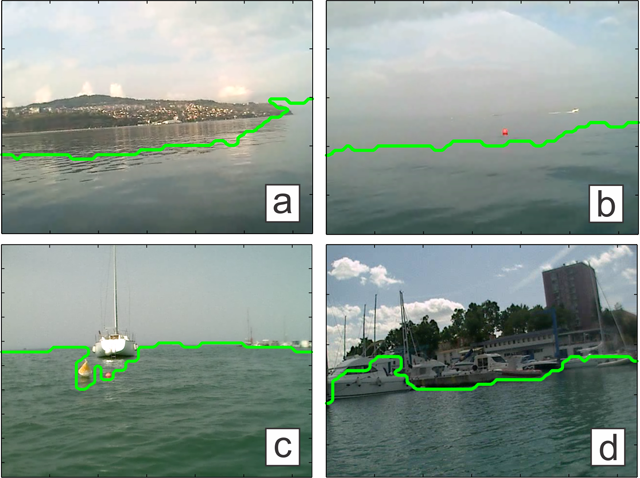}
        \caption{\label{fig:poorSegmentation} Qualitative examples of poor segmentation.}
\end{figure}

An example of conditions is which the segmentation is expected to fail is shown in the bottom-most right image of Figure~\ref{fig:goodsegmentation}. In this image, the boat is facing a low-laying sun directly, which results in a large saturated glitter on the water surface. Since the glitter occupies a large region, and is significantly different from the water, it is detected as an obstacle. Such cases could be handled by image postprocessing, but at a risk of missing true detections. Nevertheless, additional sensors like compass, IMU and sun-position model can be used to identify a detected region as a potential glitter. To offer further insights of the constraints of the proposed segmentation, we show additional failure cases in Figure~\ref{fig:poorSegmentation}. Figure~\ref{fig:poorSegmentation}a shows failure due to a strong reflection of the landmass in the sea, while Figure~\ref{fig:poorSegmentation}b shows an example of failure due to blurred transition from the sea to sky. Note that in both cases, the edge of the sea is conservatively estimated, meaning that true obstacles were not mislabelled as water, but rather portions of water were labelled as obstacle. Figure~\ref{fig:poorSegmentation}c shows an example in which several obstacles are close-by and are not detected as separate obstacles, but rather as part of the edge of water. An example of potentially dangerous mislabelling is shown in Figure~\ref{fig:poorSegmentation}d, where a part of the boat on the left is deemed visually-similar to water and is labelled as such. Note, however, that this mislabelling is corrected in the subsequent images in that video.

\subsection{Effects of the target size}

\begin{figure}
        \centering
            \includegraphics[width=6.5cm]{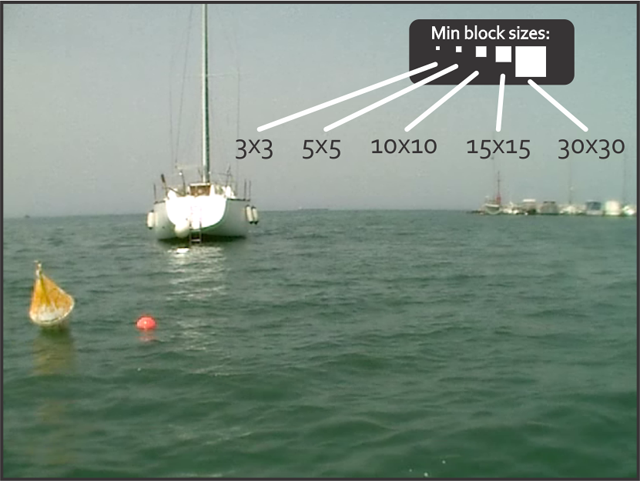}
        \caption{\label{fig:visualizationOfBlocks} Visualization of minimum thresholds used in Table~\ref{tbl:resultsSize}.}
\end{figure}

Note that all obstacles may not pose equal threat to the vessel. In fact, smaller objects are likely not people and may also likely pose little threat, since they can be run over without damaging the vessel. To inspect our results in such a context, we have compiled the $\mathrm{SSM}$ results over all videos with respect to the minimum object's size. Any object, whether in the ground truth or in detection, was ignored if its size was smaller than a predefined value. We also ignored any detected object that overlapped with the removed ground truth detection by 0.3. This last condition addresses the fact that some objects in the ground truth are slightly smaller than their detected size, which would generate an incorrect false positive if the ground truth object was removed. Figure~\ref{fig:visualizationOfBlocks} visualizes the applied thresholds, while the results are given in Table~\ref{tbl:resultsSize}.

The results show that the detection remains high over a range of small thresholds, which speaks of a level of robustness of our approach. By increasing thresholds above $10 \times 10$ the precision as well as the recall increase the probability of detecting a false positive in a given frame is drastically reduced. This means that, as the objects approach the USV and get bigger, they are increasingly reliably detected. This is also true for the sufficiently big objects that are far away from the USV. The following rule-of-thumb calculation for the big or approaching objects can be performed. Let us assume that a successful detection means any detection of a true obstacle if we detect it at least once in $N_\mathrm{buf}=3$ consecutive frames. The probability of a successful detection is therefore
\begin{equation}\label{eq:prob_fail}
    p_\mathrm{success} = 1 - (1 - \mathrm{Rec})^{N_\mathrm{buf}}.
\end{equation}
If we do not apply any thresholding, we can detect any object, regardless of its size with probability $0.988$. The probability of a false positive occurring in any image is 0.055. By applying a small ${3\times 3}$ threshold, the detection remains unchanged, but the  probability of a false positive occuring in a particular frame goes down to $0.05$. If we chose to focus only on the objects that are at least thirty by thirty pixels large, then the probability of detection goes up to $0.992$, and the probability of detecting a false positive in any frame goes down to $0.01$. It should be noted that the model in (\ref{eq:prob_fail}) assumes independence of detections over the sequence of images. While such assumptions may indeed be restrictive for temporal sequences, we still believe that the model gives a good rule-of-thumb on expected real-life obstacle detection performance of the segmentation algorithm.


\begin{table}[h!]
    \caption{\label{tbl:resultsSize} The results of the $\mathrm{SSM}_{a\times a}$ and related approaches on all video sequences with respect to the minimum object size ${a\times a}$. For reference, the results for top-performing baselines are provided for $5\times 5$, $15\times 15$ and $30\times 30$ pixels.}
    \centerline{
        \begin{tabular}{|@{\extracolsep{\fill}}l |c c c c |}
    \hline
    & $\mathrm{Prec}$ & $\mathrm{Rec}$ & $F$  & {aFP}   \\
    \hline
         $\mathrm{SSM}_{0\times0}$   & 0.885 & 0.772 & 0.819 & 0.055 \\
         $\mathrm{SSM}_{3\times3}$   & 0.898 & 0.772 & 0.825 & 0.049 \\
         $\mathrm{SSM}_{5\times5}$   & 0.898 & 0.772 & 0.825 & 0.049 \\
         $\mathrm{SSM}_{10\times10}$ & 0.898 & 0.773 & 0.826 & 0.049 \\
         $\mathrm{SSM}_{15\times15}$ & 0.896 & 0.792 & 0.837 & 0.049 \\
         $\mathrm{SSM}_{30\times30}$ & 0.924 & 0.801 & 0.846 & 0.010 \\
         \hline
         $\mathrm{GCM}_{5\times5}$   & 0.733 & 0.686 & 0.703 & 0.246 \\
         $\mathrm{GCM}_{15\times15}$ & 0.731 & 0.701 & 0.701 & 0.246 \\
         $\mathrm{GCM}_{30\times30}$ & 0.812 & 0.759 & 0.772 & 0.068 \\
         \hline
         $\mathrm{FZH}_{5\times5}$   & 0.731 & 0.523 & 0.553 & 0.082 \\
         $\mathrm{FZH}_{15\times15}$ & 0.731 & 0.536 & 0.565 & 0.082 \\
         $\mathrm{FZH}_{30\times30}$ & 0.779 & 0.577 & 0.614 & 0.038 \\
         \hline
         $\mathrm{SPX}_{5\times5}$   & 0.007 & 0.001 & 0.001 & 0.078 \\
         $\mathrm{SPX}_{15\times15}$ & 0.007 & 0.001 & 0.001 & 0.078 \\
         $\mathrm{SPX}_{30\times30}$ & 0.003 & 0.001 & 0.001 & 0.074 \\
         \hline
        \end{tabular}
    }{}
\end{table}

\section{Discussion and conclusion}\label{sec:conclusion}

A graphical model for semantic segmentation of marine scenes was presented and applied to USV obstacle-map estimation. The model exploits the fact that scenes a USV encounters may be decomposed into three dominant visually- and semantically-distinctive components, one of which is the water. The appearance is modelled by a mixture of Gaussians and accounts for the outliers by a uniform component. The geometric structure is enforced by placing weak priors over the component means. A MRF model is applied on prior and posterior pixel-label distribution to account for the interactions across neighboring pixels. An EM algorithm is derived for fitting the model to image, which affords fast convergence and efficient implementation. The proposed model directly applies straight-forward features, i.e., color channels and pixel positions and avoids potentially slow extraction of more complex features. Nevertheless, the model is general enough to be directly applied without modifications to any other features. A straightforward approach for estimation of the weak prior model was proposed, that allows learning from a small number of training images and does not require accurate annotations. Results show excellent performance compared to related segmentation approaches and exhibits improved performance in terms of segmentation accuracy as well as speed.

To evaluate the performance and analyze our algorithm, we have compiled and annotated a new real-life coastal line segmentation dataset captured from an onboard marine vehicle camera. This is the largest dataset of its kind to date and is as such another contribution to the field of robotic vision. We have studied the effects of the colorspace selection on the algorithm's performance. We conclude that the algorithm is fairly robust to this choice, but obtains best results at YCrCb and Lab colorspaces. The experimental results also show that the proposed algorithm significantly outperforms the related solutions. While the algorithm provides high detection rates at low false positives it does so with a minimal processing time. The speed comes from the fact that the algorithm can be implemented through convolutions and from the fact that it preforms robustly on small images. The results have also shown that the proposed method outperforms the related methods by a large margin in terms of robustness in the extreme conditions, when the vehicle is facing the sun, as well. To make the present paper a reproducible research and to facilitate other researchers in comparing their work to ours, the Modd dataset is made publicly available, along with all the Matlab evaluation routines, a reference Matlab implementation of the presented approach and the routines for learning the weak priors.

Note that the fast performance is of crucial importance for real-life implementations on USVs, as it allows the use in onboard embedded controllers and low-cost embedded, low-resolution cameras. Our future work will focus on two extensions of our algorithm. We will explore possibilities of porting our algorithm to such an embedded sensor. Since many modern embedded devices contain GPUs, we will also explore parallelization of our algorithm by exploiting the fact that it is based on convolution operations, which can be efficiently parallelized. Our model is fully probabilistic and as such affords a principled way for information fusion, e.g.,~\cite{Tick2013}, to improve performance. We will explore combinations with additional external sensors such as inertial sensors, cameras of other modalities and stereo systems. In particular, IMU can be used to modify the priors and soft reset parameters on-the-fly as well as estimating the position of the horizon in the images. The segmentation model can then be constrained by hard-assigning pixels above the horizon to the non-water class. Temporal constraints on segmentation can be further imposed by image-based ego-motion estimation using techniques from structure-from-motion.

\ifCLASSOPTIONcaptionsoff
  \newpage
\fi

\bibliographystyle{styles/IEEEtran}
\bibliography{\bibliographydir/bib}

\end{document}